\documentclass[11pt]{article}

\usepackage[preprint]{acl}

\usepackage{times}
\usepackage{latexsym}

\usepackage[T1]{fontenc}

\usepackage[utf8]{inputenc}

\usepackage{microtype}

\usepackage{inconsolata}

\usepackage{graphicx}

\usepackage{tcolorbox}
\tcbuselibrary{theorems}
\usepackage{spverbatim}
\usepackage{listings}
\usepackage{color}
\usepackage{fancyvrb}
\usepackage{enumitem} 

\usepackage[svgnames]{xcolor}

\usepackage{multirow}
\usepackage{booktabs}
\usepackage{makecell}

\definecolor{c_head}{RGB}{192, 227, 242}
\definecolor{c_relation}{RGB}{255, 155, 165}
\definecolor{c_tail}{RGB}{209, 195, 225}

\newcounter{colorboxctr}

\title{Wikontic: Constructing Wikidata-Aligned, Ontology-Aware \\ Knowledge Graphs with Large Language Models}

\author{
Alla Chepurova$^{1,2}$\quad\
Aydar Bulatov$^{1,2}$\quad
Mikhail Burtsev$^3$\quad
Yuri Kuratov$^{1,2}$\\\\
$^1$Cognitive AI Systems Lab, Moscow, Russia \\
$^2$Moscow Independent Research Institute of Artificial Intelligence, Moscow, Russia\\
$^3$London Institute for Mathematical Sciences, London, UK\\
\texttt{\{chepurova,bulatov,kuratov\}@cogailab.com}, \texttt{mb@lims.ac.uk}\\\\
}

\begin{document}
\maketitle
\begin{abstract}

Knowledge graphs (KGs) provide structured, verifiable grounding for large language models (LLMs), but current LLM-based systems commonly use KGs as auxiliary structures for text retrieval, leaving their intrinsic quality underexplored.
In this work, we propose \emph{Wikontic}, a multi-stage pipeline that constructs KGs from open-domain texts by extracting candidate triplets with qualifiers, enforcing Wikidata-based type and relation constraints, and normalizing entities to reduce duplication.
The resulting KGs are compact, ontology-consistent, and well-connected; on MuSiQue, the correct answer entity appears in 96\% of generated triplets.
On HotpotQA, our triplets-only setup achieves 76.0 F1, and on MuSiQue 59.8 F1, matching or surpassing several retrieval-augmented generation baselines that still require textual context. In addition, Wikontic attains state-of-the-art information-retention performance on the MINE-1 benchmark (86\%), outperforming prior KG construction methods.
Wikontic is also efficient at build time: KG construction uses less than 1,000 output tokens, about 3$\times$ fewer than AriGraph and $<$1/20 of GraphRAG.
The proposed pipeline improves the quality of the generated KG and offers a scalable solution for leveraging structured knowledge in LLMs. Wikontic is available at \url{https://github.com/screemix/Wikontic}.

\end{abstract}

\section{Introduction}

A substantial amount of the world’s knowledge exists solely in unstructured textual form, such as news, scientific articles, blogs, and posts on social networks. While large language models (LLMs) are capable of extracting insights from these data, their internal representations are latent and often unverifiable, making them susceptible to hallucinations. In contrast, knowledge graphs (KGs) record information as explicit subject-relation-object triplets, supporting verifiable queries, incremental updates, and multi-step reasoning (e.g., in answering compositional questions), making KGs a reliable complement to both LLMs and retrieval-augmented generation (RAG) systems. Therefore, creating high-quality KGs directly from raw text provides reliable, transparent knowledge that complements LLMs and RAG systems.


Extracting structured knowledge from text is a long-standing challenge in Information Retrieval. A common formulation is closed information extraction (cIE), which assumes predefined fixed sets of entities and relation predicates drawn from an existing KG and seeks to recover all triplets that conform to that schema. Classical cIE pipelines decompose the task into stages such as named entity recognition and relation classification~\cite{zeng2014relation, zhang2020minimize}. However, separating these stages leads to error accumulation and prevents the sharing of information between tasks. More recent end-to-end approaches approach extraction as a sequence-to-sequence problem, training models to directly generate triplets from text~\cite{distiawan2019neural, cabot2021rebel, josifoski2021genie} or complete missing links in KG~\cite{yao2019kg}. While this reduces error propagation, such models remain difficult to adapt to new domains, requiring costly retraining on high-quality annotated corpora that remain scarce. LLMs offer a promising alternative: their broad knowledge and strong prompting abilities enable open-domain extraction without expensive task-specific training~\cite{wang2020language, josifoski2023exploiting, chepurova2024prompt}.

In contrast to cIE, open information extraction (oIE) does not impose predefined entity and relation names, ontology constraints, and constructs KGs from scratch. This flexibility makes oIE an attractive tool for augmenting LLMs and RAG systems, with recent work demonstrating reduced inference costs and more reliable retrieval~\cite{chen2024sac, gutierrez2024hipporag, guo2024lightrag, li2024graphreader, han2024retrieval, gutierrez2025rag}. For example, AriGraph~\cite{anokhin2024arigraph} learns KGs to create long-term semantic memory, while Distill-SynthKG~\cite{choubey2024distill} integrates mixed text–triplet structures to improve question answering. However, most current oIE pipelines rely on KGs mainly as auxiliary scaffolds to structure text retrieval, rather than treating the KG itself as a high-quality knowledge resource. This perspective overlooks the potential of compact and non-redundant KGs to represent information directly, and leaves their quality and reasoning capabilities underexploited. As a result, the practical use of oIE KGs remains limited. Extracted triplets often contain heterogeneous surface forms---for example, “NYC located in USA” versus “New York City in-country United States”---fragmenting the KG into redundant or inconsistent representations. Synonymy, coreference, and predicate variation accumulate with scale, eroding the very strengths that motivate KG construction in the first place: precision, interpretability, and logical consistency.

To address this, we combine the flexibility of oIE with the structural rigor of cIE by leveraging external ontologies. Wikidata~\citep{vrande2012wikidata}, one of the largest community-maintained knowledge bases, offers rich entity classes, relation schemas, and domain-range constraints across more than 100M entities. Its breadth allows coverage from common sense to specialized domains, while its formal constraints provide principled supervision for validating LLM outputs. Yet, integrating such ontology guidance into a fully automated pipeline poses key challenges: (i) extracting candidate triplets without predefined labels, (ii) typing and disambiguating entities under ontology classes despite lexical ambiguity, and (iii) refining nodes and edges iteratively while preserving alignment.

In this paper, we address these challenges with \textbf{Wikontic}, a multi-stage framework that constructs Wikidata-aligned, ontology-aware KGs directly from texts using LLMs. Unlike prior works that apply Wikidata ontology only for evaluation or entity linking~\cite{polat2025testing}, we \emph{integrate a large-scale ontology from Wikidata directly into the information extraction pipeline}. Wikontic includes six components: (i) a curated ontology database derived from Wikidata, (ii) candidate triplet extraction with qualifiers, (iii) ontology-aware triplet refinement enforcing schema constraints, (iv) subject/object name refinement for entity deduplication, (v) KG storage, and (vi) retrieval for multi-hop question answering. Starting from unstructured text, Wikontic extracts triplets with LLMs, types, and validates them against Wikidata, and deduplicates entities to yield compact, consistent KGs.

The resulting KGs are both interpretable and effective. When used as the sole knowledge source for multi-hop QA, Wikontic achieves competitive performance with RAG and KG-based methods that rely on raw text as context~\cite{lee2024human, li2024graphreader, anokhin2024arigraph, panda2024holmes, gutierrez2025rag}. Moreover, our graphs exhibit superior coverage of salient information and strong internal connectivity between relevant nodes.
In summary, our main \textbf{contributions} are:

\noindent 1. We introduce \emph{Wikontic}, which (1) extracts candidate triplets, (2) enforces schema and ontology constraints on entity types and domain–range, and (3) performs alias-aware entity normalization and deduplication to reduce redundancy, yielding ontology-consistent KGs.

\noindent 2. We show that Wikontic's KGs are ontology-consistent, have low redundancy, strong coverage, and connectivity; on MuSiQue, the correct answer entity is present in 96\% of generated triplets.

\noindent 3. Using a KG as the sole knowledge source (no access to the original text) on multi-hop question answering, Wikontic attains 76.0 F1 on HotpotQA and 59.8 F1 on MuSiQue, matching or surpassing several RAG/KG baselines that still rely on text.

\noindent 4. Wikontic achieves state-of-the-art results on the MINE-1 benchmark, reaching 86\% information-retention score.

\noindent 5. Wikontic's KG construction uses less than 1,000 output tokens, which is \(\sim\)3$\times$ fewer than AriGraph and $<$1/20 of GraphRAG.

\section{Methods: Wikontic}

\begin{figure*}[t]
    \centering
    \includegraphics[width=0.9\textwidth]{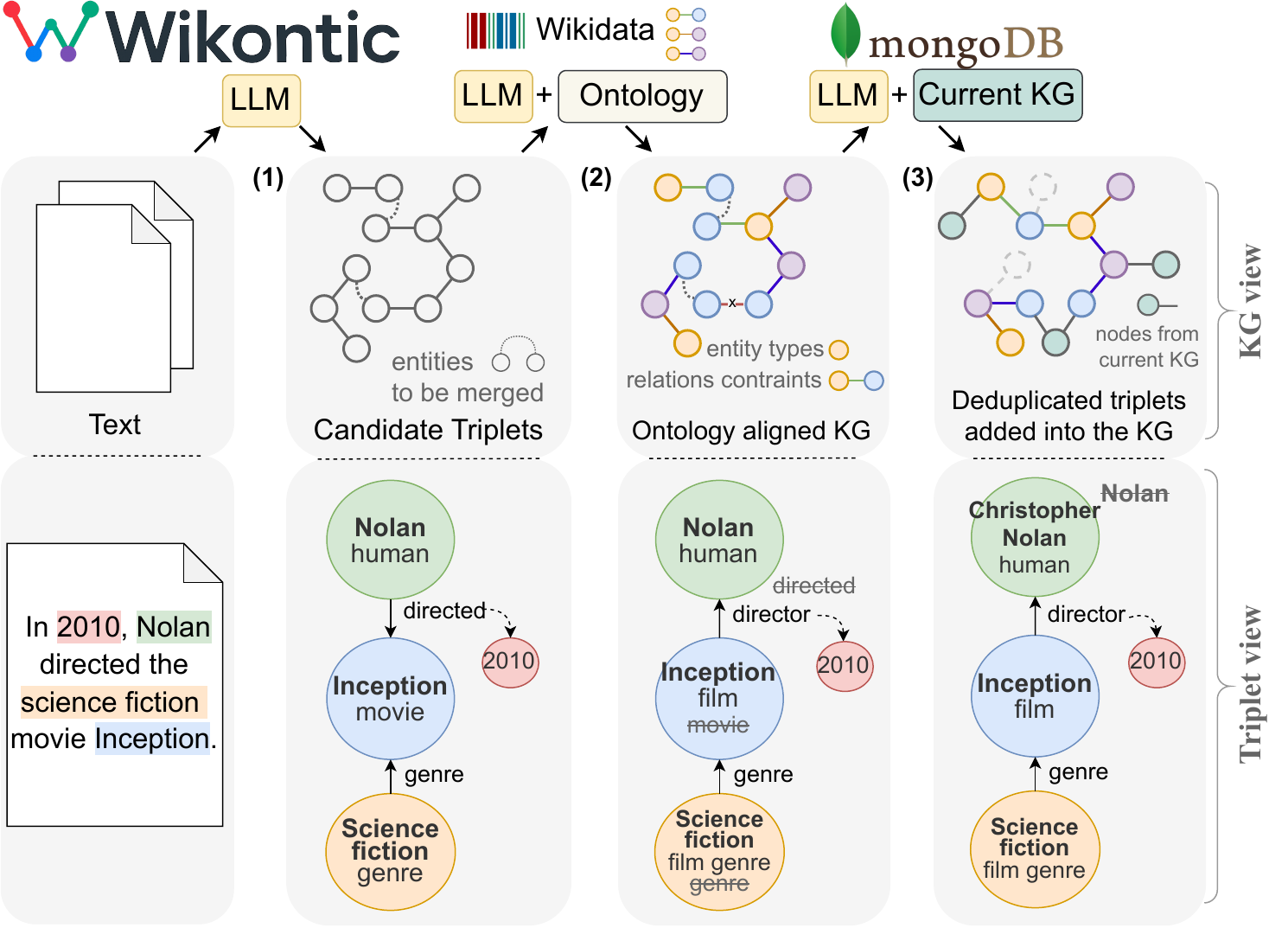}    
\caption{Overview of \emph{Wikontic}: an ontology-guided pipeline that constructs a Wikidata-aligned KG from text.
(1) An LLM extracts candidate (subject, relation, object) triplets (gray). (2) The extracted triplets are then refined using Wikidata’s ontology: entity types are assigned (colored nodes), and relations that violate ontology constraints are corrected or removed. (3) Finally, entity names are normalized, and duplicated surface forms are merged. The resulting graph is de-duplicated, ontology-consistent, and ready for downstream tasks.}
\label{fig:pipeline}
\end{figure*}


Wikontic is a multi-stage pipeline for constructing high-quality, ontology-aware KGs directly from unstructured text (Figure~\ref{fig:pipeline}). Unlike prior approaches that directly map text to graph form and often yield noisy, redundant, or inconsistent outputs, our pipeline explicitly integrates LLMs with Wikidata-derived ontological constraints, entity normalization, and iterative refinement. 

To enable triplet validation and alignment, the pipeline stores ontology rules and the current KG (Section~\ref{sec:onto_database}). The triplet extraction pipeline (Section~\ref{sec:ie}) consists of three main stages: (i) triplet candidate extraction with contextual metadata, (ii) ontology-aware triplet refinement, and (iii) entity normalization and deduplication. These stages aim to enforce structural validity and reduce redundancy, to produce a cleaner and more semantically coherent KG that can replace raw text in RAG for multi-hop QA tasks (Section~\ref{sec:qa}).

\subsection{Ontology and KG Databases}
\label{sec:onto_database}
We built a custom \textbf{ontology schema database} derived from Wikidata. The schema database includes properties (i.e., relations) and their compatible entity types. Properties required solely for linking external data (e.g., multimedia or external identifiers) were excluded, leaving 2,464 factual properties with suitable datatypes (e.g., WikibaseItem, Quantity, Point in time).

For each property, we retrieved subject and object type constraints from Wikidata (e.g., Q21503250 for subject, Q21510865 for object). These constraints define type compatibility rules, specifying which entity classes a relation can logically connect and thereby guiding ontology-consistent triplet construction.

To support constraint generalization, we recursively expanded entity types using 'instance of' (P31) and 'subclass of' (P279) relations, building full taxonomies from each type up to the root. Such a hierarchy is essential because relation constraints are defined at different levels of abstraction. For example, a relation may allow connections between instances of the broader class 'audiovisual work', even if the entity is typed more specifically as 'film'.  By propagating the allowed properties of each parent type downwards to its children entity type, we ensure that entities can still be matched to valid relations whenever the constraint applies to its parent class.

We collected labels and aliases for all entity types and relations. Dense retrieval indexes for relation and entity type names, as well as their respective aliases, support semantic search. These indexes allow us to semantically align relation and entity type names from extracted triplets with Wikidata definitions, even when surface forms differ. 

\textbf{The KG database} stores triplets, canonical entity names, and aliases. A dense retrieval index over aliases supports efficient linking and deduplication. As extraction proceeds, new entities are inserted with canonical labels and aliases, keeping the KG compact yet incrementally extensible.

Dense retrieval indexes used in both databases were built with Contriever embeddings~\cite{contriever} and Atlas MongoDB vector search\footnote{\url{https://www.mongodb.com/products/platform/atlas-vector-search}}. MongoDB’s hybrid support for structured queries and dense retrieval enables both efficient graph and semantic search.

\subsection{Ontology-aware Triplet Extraction}
\label{sec:ie}
\paragraph{Stage 1: Candidate Triplet Extraction.}
We extract factual triplet candidates from unstructured text with an LLM, capturing subject-relation-object triplets, along with contextual qualifiers that enhance the semantic meaning of the triplet. LLM is prompted with instruction and in-context examples to extract triplets that include entity types for both the subject and object, as well as additional metadata that mirrors the structure of Wikidata qualifiers. These qualifiers are essential because they capture contextual information such as time, location, or conditions. While such details usually cannot be expressed as standalone facts, they are critical for preserving factual precision and avoiding loss of accurate knowledge during knowledge extraction. 


For instance, given the text \textit{"In 2010, Christopher Nolan directed the science fiction movie Inception"}, the extracted triplet would be:
\texttt{(Nolan, directed, Inception)}
with entities types \texttt{(human, film)} and the qualifier: \texttt{\{point in time: 2010\}}. Further details and examples are provided in Appendix~\ref{sec:prompts}.

However, LLM outputs may be semantically redundant or structurally inconsistent. The entity and relation names may not align with existing entities and relations already present in the KG. For example, an LLM might extract "Nolan" as the subject from one input text and extract "Christopher Nolan" from the other one; or relations like "directed" vs. "director" might appear in different grammatical forms or inverse directions in different input texts (see Figure~\ref{fig:pipeline}, bottom). Without additional correction, these inconsistencies lead to entity duplication and increased KG size, which degrades both storage efficiency and downstream reasoning. Thus, to improve consistency and reduce redundancy, the next steps of the pipeline validate extracted triplets using Wikidata’s ontology and normalize entity names.



\paragraph{Stage 2: Ontology-aware Refinement.}
At this stage, each candidate triplet is refined using the schema and constraints of the Wikidata ontology:

Entity typing: For both subject and object, we retrieve the top-10 candidate types from the dense retrieval index. The LLM then selects the most plausible type. We then add supertypes from the taxonomy to ensure coverage when constraints are defined at higher abstraction levels.

Relation validation: Using Wikidata constraints, we identify all relations that can legally connect selected subject and object types, including inverse combinations of subject and object types (e.g., directed vs. director). These candidate relations are ranked by cosine similarity to the originally extracted relation.

Triplet backbone reconstruction: The text, triplet, and valid relations are passed to the LLM, which selects the most plausible ontology-valid configuration, yielding a refined triplet backbone.



This stage enforces structural validity, semantic alignment, and consistency with Wikidata’s ontology. A worked example is shown in Figure~\ref{fig:backbone} (Appendix~\ref{sec:ie_examples}).

\paragraph{Stage 3: Entity Normalization and Alias-aware Deduplication.}
While the focus of the previous step is validating triplet structure and semantics, this step aligns entity names to a unified vocabulary of existing KG entries to reduce duplication and ensure consistency of the constructed KG.


For each refined triplet, we link its subject and object names to existing entities in the KG that share the same entity type or a compatible parent type from the taxonomy. Using precomputed embeddings of entity aliases from the KG, we retrieve top-10 candidates and rank them by cosine similarity to the surface forms of the extracted mentions. The top-10 candidates, together with their types, are passed to the LLM to determine whether the extracted entity is synonymous with one of the existing entries. On a match, we replace the mention with the canonical KG label and store the surface form as an alias; otherwise, we preserve a new entity and add its surface form to the alias collection.

This step aims to ensure that the resulting KG is compact by avoiding redundant entities with different surface forms and evolving by supporting incremental updates with the discovery of new entities. A detailed example for the second step is provided in Figure~\ref{fig:names} in Appendix~\ref{sec:ie_examples}.

We implement \textbf{a final ontology verification step} to ensure that extracted triplets comply with the structural and semantic constraints of the target KG. A triplet is verified if (i) its subject and object types, together with the relation, are defined in the ontology, and (ii) the relation’s domain and range constraints are satisfied. Triplets that fail these checks are flagged as ontology-misaligned but retained, as they remain linked to the main KG through the entity name refinement step. Preserving them enables computation of an ontology alignment score and provides interpretable cues for identifying or revising schema-inconsistent facts.

\subsection{Retrieval for QA}
\label{sec:qa}

We address multi-hop question answering with the constructed KG via an iterative retrieval that decomposes the question into subquestions, grounding each step in the retrieved KG context.

Given a question, the LLM decomposes it into the first 1-hop subquestion. For each subquestion the LLM (1) identifies explicitly mentioned or potentially relevant entities; (2) links extracted entities to KG nodes and selects those most relevant for the current step; (3) given the retrieved subgraph formed by the neighborhood of the selected entities, generates an answer to the subquestion; (4) conditioned on the previous answer, the LLM formulates the next subquestion. This iterative process continues for up to five subquestions, after which the LLM produces the final answer.
Implementation details and prompt templates are provided in Appendix~\ref{sec:qa_prompts}.


\subsection{Evaluation}

Existing benchmarks for triplet extraction suffer from substantial limitations: annotated closed IE datasets are small-scale, noisy, and often incomplete~\cite{josifoski2023exploiting, josifoski2021genie, huguet-cabot-navigli-2021-rebel}, while open IE corpora are difficult to align with real-world KGs and provide unreliable ground truth for evaluation~\cite{stanovsky2018supervised}. Constructing high-quality datasets is costly, as annotators must not only identify all explicit and implicit entity and relation mentions but also align them to the complex schemas of large KGs such as Wikidata, which contain thousands of entity types and relations. Recently, the MINE benchmark~\cite{mo2025kggen} was introduced to address some of these issues by evaluating KGs through \textit{information retention} rather than exact triplet-level supervision. We evaluated Wikontic on the MINE-1 task and observed much higher information retention performance. However, while MINE provides a useful and scalable proxy for KG quality, it does not include complete ground-truth triplets. Therefore, it cannot measure classical precision or recall and instead captures only the degree to which a constructed KG preserves factual information.

To further address this, we adopt an alternative evaluation strategy. 
We measure KG quality through (1) structural compactness (i.e., non-redundancy and deduplication) and (2) performance in downstream multi-hop QA. In this setup, the LLM must answer factual questions in a text-free setting using only the constructed KG, without access to the original source texts, unlike retrieval-augmented methods such as HippoRAG~\cite{gutierrez2024hipporag}, AriGraph~\cite{anokhin2024arigraph}, and Holmes~\cite{panda2024holmes}. This design makes QA a functional proxy for two key properties of an extracted KG: (a) factual correctness, since noisy or invalid triplets directly impede correct answers, and (b) coverage and completeness, since incomplete graphs restrict multi-hop reasoning. 
Despite existing advances in aligning KGs with LLMs and adapting for QA~\cite{han2022self, dai2025large, sui2024can, pan2024unifying}, we deliberately refrain from training additional models to estimate the KG quality itself.

To compare predicted answers with ground truth, we apply a normalization procedure that lowercases all strings and removes punctuation. To account for lexical variation, we further expand entity matching using the alias mappings stored in the KG. If the model’s predicted answer matches any canonical entity or one of its aliases, the corresponding alias set is treated as the set of valid candidate answers. 

We perform evaluations on KGs extracted using different LLMs: \texttt{gpt-4.1}, \texttt{gpt-4.1-mini}, \texttt{gpt-4o-mini}\footnote{\url{https://platform.openai.com/docs/models}}, and \texttt{Llama-3.3-70b-Instruct}\footnote{\url{https://huggingface.co/meta-llama/Llama-3.3-70B-Instruct}}, and assess the quality of the resulting KG on two multi-hop QA datasets: MuSiQue~\cite{trivedi2022musique} and HotpotQA~\cite{yang2018hotpotqa}.  We used the same questions and candidate passages, including both supporting and distractor passages, used in HippoRAG~\cite{gutierrez2024hipporag} and AriGraph~\cite{anokhin2024arigraph} to compare the results with existing methods.

\begin{figure*}[ht!]
    \centering
    \includegraphics[width=0.85\textwidth]{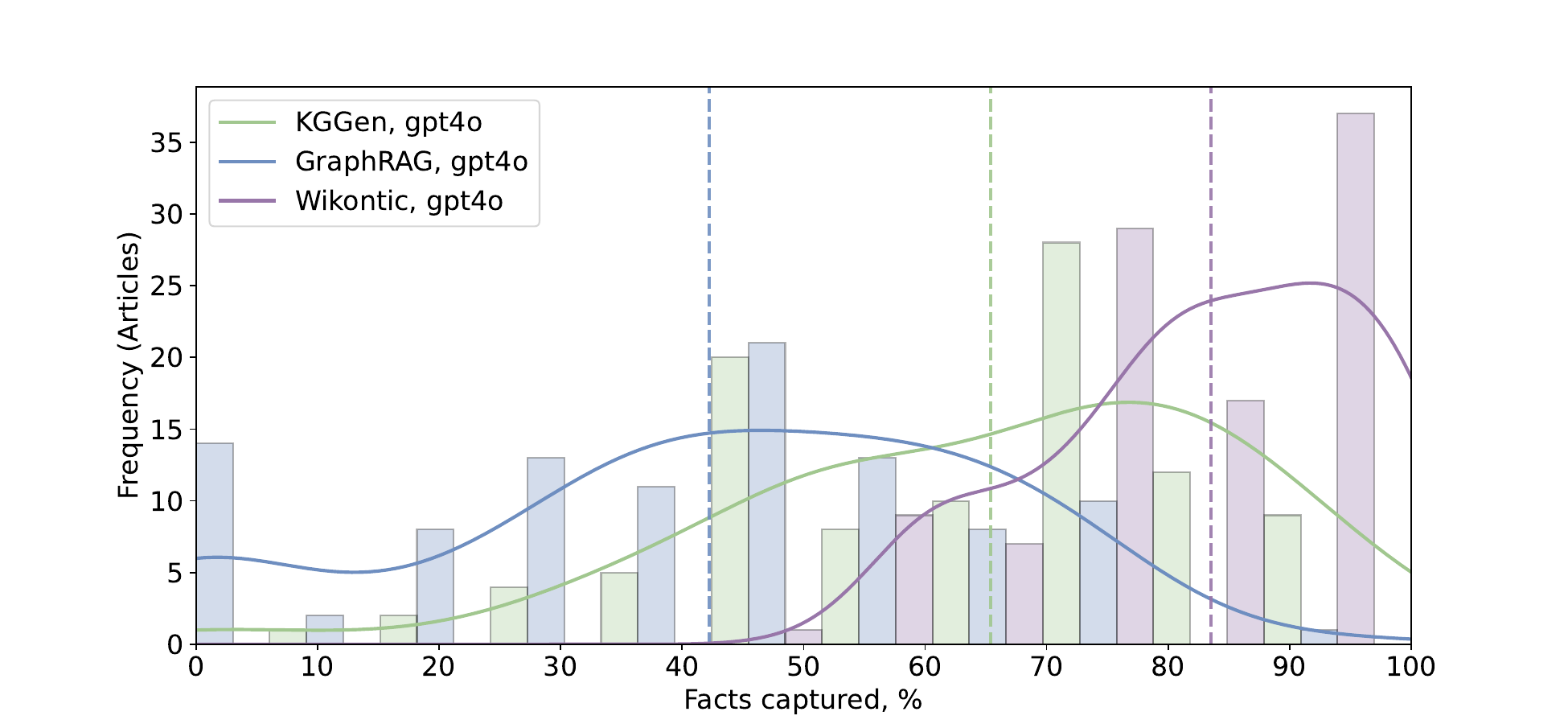}    
\caption{Distribution of MINE-1 scores across 100 articles for GraphRAG, KGGen, and Wikontic.
Dotted vertical lines are averaged scores. Wikontic scored 84\% on average, substantially 
outperforming GraphRAG 47.80\% and KGGen 66\%.}
\label{fig:mine_dist}
\end{figure*}

\section{Results}
\subsection{Evaluations on MINE-1}

We evaluated Wikontic on the MINE-1 benchmark, which measures how much factual information from the source text is retained in the constructed KGs using an LLM-as-a-judge protocol from the original study~\cite{mo2025kggen}. Figure~\ref{fig:mine_dist} displays the retention scores distribution in articles of MINE-1 for KGGen, GraphRAG, and Wikontic. Table~\ref{tab:mine} demonstrates the results for both KGGen and Wikontic with different LLM backbones. Wikontic consistently outperforms KGGen, reaching 84\% with \texttt{gpt-4o} and 86\% with \texttt{gpt-4.1-mini}, compared to KGGen’s best score of 73\% (Claude Sonnet 3.5). These results demonstrate that Wikontic effectively preserves factual information during the construction of the KG.

\begin{table}[ht] 
\centering
\small
\renewcommand{\arraystretch}{1.2} 
\setlength{\tabcolsep}{3pt}
\resizebox{0.85\linewidth}{!}{%
\begin{tabular}{lr}
\hline
\textbf{Method} & \textbf{MINE-1 Score (\%)} \\
\hline
KGGen, Claude Sonnet 3.5 & 73 \\
KGGen, GPT-4o & 66 \\
KGGen, Gemini 2.0 Flash & 44 \\
GraphRAG, gpt4o & 48 \\
Wikontic, gpt4o & \textbf{84} \\
Wikontic, gpt4.1-mini & \textbf{86} \\
\hline
\end{tabular}
}
\caption{MINE-1 information-retention scores for KGGen, GraphRAG, and Wikontic. Wikontic achieves the highest retention performance across all evaluated LLMs.}
\label{tab:mine}
\end{table}

\subsection{Graph Quality Analysis}

We aim to comprehensively evaluate the structure, information content, and usability of knowledge graphs created by Wikontic in a challenging information extraction setting. Given the limited availability of approaches that assess the KG quality directly, we use a proxy evaluation methodology based on the MuSiQue QA dataset to examine how effectively the knowledge is represented in the resulting KG.

A KG can be formally represented as $\mathcal{G} = (\mathcal{T}, \mathcal{E}, \mathcal{R})$, where $\mathcal{E}$ is the set of entities $e$, $\mathcal{R}$ is the set of relations $r$ and $\mathcal{T}$ is the set of triplets: $\mathcal{T} \in \mathcal{E} \times \mathcal{R} \times \mathcal{E}$. For efficient knowledge storage and retrieval, the KG should satisfy the \textit{size}, \textit{density}, and \textit{diversity} desiderata, which can be directly evaluated using graph statistics (Table~\ref{tab:graph_statistics}).

\begin{table}[htb!]
\resizebox{\columnwidth}{!}{%
\setlength{\tabcolsep}{4pt}
\begin{tabular}{lccccc}

\hline
\textbf{Method} &  |$\mathcal{E}$| &  |$\mathcal{R}$| & \makecell{Avg. $e$ \\ degree} & \makecell{Unique\\ $e$ per $r$} & \makecell{$r$ diversity\\ per $2\times e$} \\
\hline
HippoRAG & 234.9 & 130.1 & 4.0 & 1.8 & 1.1 \\
AriGraph & 228.0 & 115.6 & 3.9 & 2.0 & 1.01 \\
Wikontic (1-3) & 248.8 & 104.8 & 4.3 & 2.5 & 1.03 \\
\makecell[l]{\hspace{0.3em}w/o ontology (2)} & 232.4 & 106.7 & 4.4 & 2.6 & 1.06 \\
\makecell[l]{\hspace{0.3em}w/o ontology (2) and\\ \hspace{1.2em} normalization (3)} & 273.0 & 140.9 & 4.2 & 2.3 & 1.09 \\
\makecell[l]{\hspace{0.3em}w/o ontology-misaligned\\ \hspace{1.2em} triplets} & 239.9 & 99.5 & 4.3 & 2.6 & 1.0 \\
\hline
\end{tabular}
}
\caption{
KGs structural statistics for MuSiQue QA corpus: the number of unique entities ($|\mathcal{E}|$) and relations ($|\mathcal{R}|$), average entity degree (Avg. $e$ degree), the number of unique entities per relation (Unique $e$ per $r$) and the average relation diversity per two entities ($r$ diversity per $2 \times e$).
}
\label{tab:graph_statistics}
\end{table}

Graph size is directly connected to the number of stored facts, and can be represented by the average number of entities and relations per MuSiQue sample, denoted by $|\mathcal{E}|$ and $|\mathcal{R}|$. Wikontic without ontology (Stage 2) and normalization (Stage 3) yields diverse KGs with the highest number of unique entities and relations, followed by HippoRAG. However, the sheer volume of relations does not necessarily make retrieval more informative. Each relation should also be well-represented across various unique entities to ensure that relation names are standardized and meaningful. This property is reflected by the average number of unique entities per relation, which is significantly higher in refinement-augmented Wikontic versions.

\begin{figure}
    \centering
    \includegraphics[width=.93\linewidth]{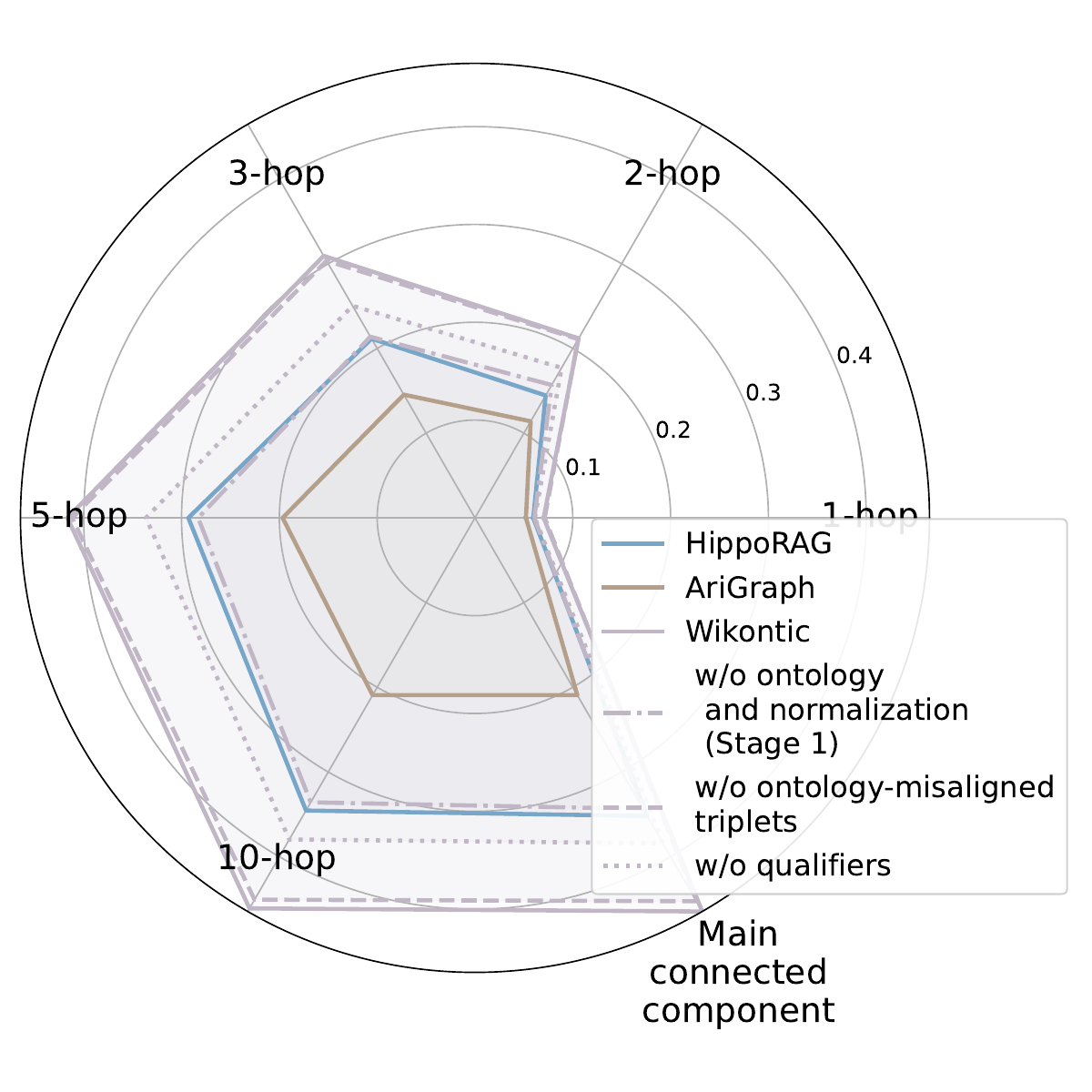}
    \vspace{-13pt}
    \caption{Wikontic produces the most dense KGs for MuSiQue questions. For each question, subgraphs are constructed around its entities, and their sizes are reported relative to the full KG. The figure shows the relative sizes of 1-- to 10-hop neighborhoods and the entire connected component containing the question, defined as all nodes reachable from any question node.}
    \label{fig:neighborhood-size}
\end{figure}

High KG connectivity, represented by the average entity degree, ensures efficient retrieval, especially with limited search depth. Entity normalization (Stage 3) is a key to building KGs with the highest density among the compared methods. The relation diversity, or the number of unique relations per two entities, captures a variety of information stored in the KG. Here, Wikontic versions with relaxed ontology constraints remain the most diverse.

In a dense graph, important nodes should have many neighbors. We select the entities from MuSiQue questions and assess their neighborhood in graphs built by various methods (Figure~\ref{fig:neighborhood-size}). The largest possible neighborhood is the main connected component containing the given entities. In KGs built by the full Wikontic pipeline, each neighborhood contains the greatest number of entities, again underscoring strong KG connectivity and the importance of ontology. 

\begin{table}[hb!]
\resizebox{\columnwidth}{!}{%
\setlength{\tabcolsep}{4pt}
\begin{tabular}{lccccc}
\hline
\multirow{2}{*}{\textbf{Method}}  &  \multicolumn{3}{c}{Contains Answer (\%)} & \multirow{2}{*}{\makecell{Ontology\\Entailment (\%)}} \\
& Total & 5-hop & 10-hop &  \\
\hline
HippoRAG & 96.3\tiny{$\pm$2.1} & \textbf{67.5}\tiny{$\pm$5.3} & 68.8\tiny{$\pm$5.2} & - \\
AriGraph & 79.9\tiny{$\pm$4.5} & 40.0\tiny{$\pm$5.5} & 41.3\tiny{$\pm$5.5} & - \\
Wikontic & 96.2\tiny{$\pm$2.1} & 66.3\tiny{$\pm$5.3} & 68.8\tiny{$\pm$5.2} & 96.5 \\
\makecell[l]{\hspace{0.6em}w/o ontology (2)} & \textbf{97.5}\tiny{$\pm$1.7} & 66.3\tiny{$\pm$5.3} & \textbf{70.0}\tiny{$\pm$5.2} & 15.2 \\
\makecell[l]{\hspace{0.6em}w/o ontology (2) and\\ \hspace{1.2em} normalization (3)} & 96.2\tiny{$\pm$2.1} & 65.0\tiny{$\pm$5.4} & 68.8\tiny{$\pm$5.2} & 12.4 \\
\makecell[l]{\hspace{0.6em}w/o ontology-misaligned\\ \hspace{1.2em} triplets} & 93.8\tiny{$\pm$2.7} & 63.8\tiny{$\pm$5.4} & 66.3\tiny{$\pm$5.3} & \textbf{100.0} \\
\makecell[l]{\hspace{0.6em}w/o qualifiers} & 85.0\tiny{$\pm$4.0} & 51.3\tiny{$\pm$5.7} & 53.8\tiny{$\pm$5.6} & 96.5 \\
\hline
\end{tabular}
}
\caption{
Knowledge coverage of graphs built using different extraction methods on the MuSiQue dataset with mini-sized models.
(Left) Percentage of cases where the correct answer to a question appears in the full constructed graph or within the 5-and 10-hop neighborhoods of the question nodes. Wikontic pipelines provide the best answer coverage while maintaining high ontology agreement.
(Right) Average percentage of triplets in each sample that are entailed by the Wikidata ontology.
}
\label{tab:compact_graph_comparison}
\end{table}

\begin{table*}[ht]
\vspace{-5pt}
\centering
\small
\begin{tabular}{lllll}
\toprule
\textbf{Method} & \multicolumn{2}{c}{\textbf{MuSiQue}} & \multicolumn{2}{c}{\textbf{HotpotQA}} \\
\cmidrule(lr){2-3} \cmidrule(lr){4-5}
 & \textbf{EM} & \textbf{F1} & \textbf{EM} & \textbf{F1} \\
\midrule
Wikontic, gpt4.1 & 46.8\tiny{$\pm{0.8}$}& \textbf{59.8}\tiny{$\pm{0.3}$} & 64.5\tiny{$\pm{0.4}$} & 76.0\tiny{$\pm{0.4}$} \\
Wikontic, gpt4.1-mini & 42.6\tiny{$\pm{0.7}$} & 55.9\tiny{$\pm{0.3}$}  & 59.7\tiny{$\pm{0.6}$}  & 71.7\tiny{$\pm{0.8}$}  \\
Wikontic, gpt4o-mini & 42.1\tiny{$\pm{0.1}$} & 53.3\tiny{$\pm{0.1}$} & 53.7\tiny{$\pm{1.2}$} & 65.8\tiny{$\pm{1.0}$} \\
\midrule
Full context, gpt4 & 33.5 & 42.7 & 53.0 & 68.4\\
Supporting facts, gpt4 & 45.0 & 56.0 & 57.0 & 73.8\\
ReadAgent~\cite{lee2024human}, gpt4 & 35.0 & 45.1 & 48.9 & 62.0\\
GraphReader~\cite{li2024graphreader}, gpt4 & 38.0 & 47.4 & 55.0 & 70.0  \\
GraphRAG~\cite{edge2024local}, gpt4o-mini & 40.0 & 53.5 & 58.7 & 63.3  \\
AriGraph~\cite{anokhin2024arigraph}, gpt4o-mini & 36.5 & 47.9 & 60.0 & 68.0 \\
AriGraph~\cite{anokhin2024arigraph}, gpt4 & 45.0 & 57.0 & \textbf{68.0} & 74.7 \\
HOLMES~\cite{panda2024holmes}, gpt4 & \textbf{48.0} & 58.0 & 66.0 & \textbf{78.0} \\
\midrule
Wikontic, Llama 3.3 & \textbf{37.7}\tiny{$\pm{0.6}$} & \textbf{49.7}\tiny{$\pm{0.4}$} & 55.1\tiny{$\pm{0.5}$} & 67.4\tiny{$\pm{0.5}$} \\
HippoRAG v2~\cite{gutierrez2025rag}, Llama 3.3 & 37.2 & 48.6 & \textbf{62.7} & \textbf{75.5} \\ 
\bottomrule
\end{tabular}
\caption{Exact Match (EM) and F1 scores on the MuSiQue and HotpotQA. Wikontic operates solely on KG triplets without accessing the source text, yet achieves performance comparable to or even exceeding both raw-text baselines (Full Context, Supporting Facts) and retrieval-augmented KG approaches that still rely on source text access.}
\label{tab:qa_performance}
\vspace{-5pt}
\end{table*}

\subsection{Answer Coverage}

In the task of question answering, the primary purpose of the extracted KG is to extract as much relevant information from the context as possible. Ideally, the resulting graph should contain all entities mentioned in the question as well as the answer to the question. In multi-hop question answering, entities in question will unlikely be in the same context as the answer entity, meaning they may not be direct neighbors in the resulting KG. However, in a KG with sufficient coverage, there has to be a reasonably short path connecting these two entities. 

To measure the overall coverage of various KG construction methods, we estimate whether the answer to the question is present in the KG as an entity and whether the path from the question to the answer exists in the graph. Due to differences in pipelines, we cast all entity and relation names to lowercase and remove punctuation to standardize their format. To account for possible differences in entity naming, we consider two entities matching if one is a substring of the other. 

Table~\ref{tab:compact_graph_comparison} presents coverage and size metrics for KGs built by our pipeline, AriGraph, and HippoRAG on the MuSiQue dataset. Since baseline KGs are available only for the gpt4o-mini model and 80 common test samples, we use the same configuration to ensure a fair comparison.
We estimated the standard deviation using bootstrapping.
"Contains Answer" represents the percentage of cases when the answer entity is present in the neighborhood of entities from the MuSiQue question. We ablate Wikontic by removing one or multiple pipeline steps at a time. "Ontology Entailment" is the percentage of triplets where subject, object, and relation match the ontology.


Both Wikontic and HippoRAG achieve over 96\% answer coverage, surpassing AriGraph’s 79.9\%. Notably, only 3.5\% of triples in Wikontic are ontology-misaligned, confirming that the generated knowledge is largely schema-consistent; only a small portion of produced triplets requires correction to fully satisfy ontology constraints.
Without ontology constraints, Wikontic reaches the highest answer coverage (97.5\%), avoiding mismatches between Wikidata and MuSiQue entities and making it particularly effective for open-domain QA. When ontology consistency is required, the Wikontic variant that excludes ontology-misaligned triplets (100\% ontology entailment) attains a competitive answer coverage of 93.8\%.

These findings suggest that Wikontic variants provide the best solution both with and without the ontology. Ontology and misaligned triplets exclusion help to achieve the most standardized graph and strong answer coverage by thorough triplet refinement and deduplication strategies. Wikontic achieves the best coverage of information essential for question answering, while maintaining the highest connectivity levels, crucial for enabling graph search.

\subsection{Computational Efficiency}




We evaluated the computational efficiency of different KG-construction methods by counting the number of input (prompt) and output (completion) tokens required to build a KG from a single paragraph in the MuSiQue dataset. Table~\ref{tab:graph-comparison} presents the estimated token-based costs for Wikontic, AriGraph, and GraphRAG, based on publicly available data and original implementations.

\begin{table}[ht] 
\centering
\renewcommand{\arraystretch}{1.2} 
\setlength{\tabcolsep}{3pt}
\resizebox{0.85\linewidth}{!}{%
\begin{tabular}{lrrr}
\hline
\textbf{Tokens} & \textbf{Wikontic} & \textbf{AriGraph} & \textbf{GraphRAG} \\
\hline
Prompt   & 12,687           & 11,000  & 115,000 \\
Completion & 881          & 2,500   & 20,000  \\
\hline
\end{tabular}
}
\caption{Mean token efficiency for KG construction per text paragraph of Wikontic compared with AriGraph and GraphRAG on the MuSiQue dataset.}
\label{tab:graph-comparison}
\end{table}

A key indicator of computational cost is the number of \textit{completion tokens}, which are typically around 3-5 times more expensive\footnote{\url{https://claude.com/platform/api/}}\textsuperscript{,}\footnote{\url{https://openai.com/api/pricing/}} and computationally intensive than input tokens~\cite{zhou2024survey}. Under this metric, Wikontic is more efficient, producing KGs with roughly three times fewer output tokens than AriGraph (881 vs 2,500) and about twenty times fewer than GraphRAG (881 vs 20,000). This shows that Wikontic achieves comparable KG construction quality while using significantly fewer output tokens.



\subsection{Performance on QA Tasks}

Table~\ref{tab:qa_performance} reports Exact Match (EM) and F1 scores on the MuSiQue and HotpotQA datasets. Unlike retrieval-augmented approaches such as HippoRAG and AriGraph, which use KGs primarily to retrieve and process relevant text passages, our method performs reasoning directly over structured triplets without accessing the original documents.

Despite this constraint, using our triplet-only contexts for \texttt{gpt-4.1}, it achieves strong results, reaching 64.5~EM and 76.0~F1 on HotpotQA and 46.8~EM and 59.8~F1 on MuSiQue. These scores surpass several retrieval-based methods, including ReadAgent and GraphReader, and are comparable to more resource-intensive systems such as AriGraph and HOLMES that rely on richer textual context. This demonstrates that complete and well-structured symbolic representations of KGs can serve as a sufficient and reliable information source for multi-hop reasoning.

\subsection{Ablation Study}

\begin{table}[ht] 
\centering
\small
\renewcommand{\arraystretch}{1.2} 
\setlength{\tabcolsep}{3pt}
\begin{tabular}{lrr}
\hline
\textbf{Method variant} & \textbf{EM} & \textbf{F1} \\
\hline
Wikontic (gpt4.1-mini)  & 42.6\tiny{$\pm{0.7}$} & 55.9\tiny{$\pm{0.3}$} \\
\hline
\hspace{0.9em}w/o qualifiers & 23.9\tiny{$\pm{0.2}$}& 39.4\tiny{$\pm{0.4}$} \\
\hspace{0.9em}w/o aliases & 36.5\tiny{$\pm{0.8}$}& 50.0\tiny{$\pm{1.4}$} \\
\hspace{0.9em}w/o ontology (2)& 36.3\tiny{$\pm{1.2}$}& 48.8\tiny{$\pm{1.0}$} \\
\makecell[l]{\hspace{0.9em}w/o ontology (2) and \\ \hfill normalization (3)} & 27.0\tiny{$\pm{1.2}$}& 36.9\tiny{$\pm{2.2}$} \\
\hspace{0.9em}Single-step QA & 31.3\tiny{$\pm{0.6}$}& 43.4\tiny{$\pm{0.6}$} \\
\hline
\end{tabular}
\caption{Ablations of the Wikontic pipeline on MuSiQue.
“Single-step QA” omits iterative subquestion reasoning. Removing ontology or entity normalization yields the largest drop, highlighting their importance for accurate reasoning over the constructed KG.}
\vspace{-4pt}
\label{tab:ablation}
\end{table}

To evaluate the contribution of individual Wikontic components, we conducted ablations (Table~\ref{tab:ablation}). Removing qualifiers leads to a substantial performance drop (–15.9 EM, –15.7 F1) on MuSiQue, indicating that qualifier information is essential for capturing fine-grained relational context. Excluding aliases (introduced at Stage 3) moderately decreases performance, confirming that alias expansion improves entity matching in question answering. Eliminating ontology integration (Stage~2) reduces both EM and F1, demonstrating the importance of type and schema constraints for consistent KG construction. When both ontology and entity normalization are removed (Stages 2 and 3), performance degrades most severely. The single-step QA variant also performs significantly worse, confirming that multi-hop question decomposition is essential for effective reasoning over the constructed KG. Overall, these findings show that each component contributes meaningfully to Wikontic’s performance, with ontology-guided refinement and iterative retrieval being the most critical for downstream reasoning over the constructed KG.

\section{Conclusions}

We introduced Wikontic, a fully automated pipeline that uses LLMs to construct KGs from unstructured text. The pipeline produces compact, internally consistent graphs by aligning extracted triplets with the Wikidata ontology and deduplicating entities and relations. 
While using KGs as the sole knowledge source for multi-hop question answering, Wikontic achieves competitive performance with retrieval-augmented and KG-based baselines that still rely on source texts. Thus, ontology-guided KG construction is a viable alternative to passage-level retrieval. On MuSiQue, Wikontic includes 38–45 more unique entities than HippoRAG and AriGraph and contains the correct answer entity in 97.5\% of cases. Within a 10-hop subgraph, it maintains 70\% answer coverage and denser local connectivity. For QA, it attains 64.5 EM / 76.0 F1 on HotpotQA and 46.8 EM / 59.8 F1 on MuSiQue, outperforming text-reliant systems (ReadAgent, GraphReader) and approaching larger text-dependent methods (AriGraph, HOLMES). Moreover, Wikontic achieves state-of-the-art results on the MINE-1 benchmark, achieving 84–86\% information-retention scores and surpassing GraphRAG and KGGen. These findings indicate that the proposed pipeline preserves a substantial amount of factual information from source texts. 




Our approach is also token-efficient: during KG construction, Wikontic uses under 1{,}000 output tokens, about $3\times$ fewer than AriGraph and $1/20$ of GraphRAG, while preserving accuracy. Moreover, only 3.5\% of extracted triplets are flagged as ontology-misaligned, indicating that nearly all generated knowledge is schema-consistent, minimizing the need for manual correction and significantly reducing annotation overhead.
Beyond efficiency, the pipeline is adaptable: it can operate without an ontology or integrate domain-specific ontologies, enabling applications across specialized domains. Moreover, as LLMs increasingly serve as data generators, Wikontic provides a principled foundation for producing verified, structured KG data suitable for fine-tuning smaller task-specific models.

Our findings show that ontology-aware KG construction enables scalable, interpretable, and verifiable transformation of unstructured text into structured knowledge, bridging symbolic reasoning and generative language modeling.

\section*{Limitations}
Our experiments are restricted to proprietary OpenAI models (GPT-4.1, GPT-4.1-mini, GPT-4o-mini) and the open-source Llama 3.3-70B. 
Token efficiency is measured as model-generated tokens during KG construction. This metric reflects provider billing but not end-to-end latency or throughput.
Every stage of the pipeline currently uses LLMs with instructions and in-context examples prompting. Because the pipeline now yields its own annotated data, several stages could be replaced in future work by smaller, task-specific models fine-tuned on this data, thereby improving efficiency and lowering computational cost.

In the current work, we focus only on Wikidata due to its size, quality, and rich ontology. However, the proposed pipeline is flexible and can be adapted to any domain and ontology with a matching format of triplet constraints and entity types. 

\bibliography{custom}

\appendix

\clearpage
\section{Appendix}

\subsection{Reproducibility}
\label{sec:reproducibility}
We release\footnote{\url{https://github.com/screemix/Wikontic}}: (1) prompts and source code for each pipeline stage; (2) scripts to build Wikidata-derived ontology used for ontology-aware stages; (3) the KG-only multi-hop QA component.

We reported metrics averaged across multiple runs (Tables~\ref{tab:qa_performance},~\ref{tab:ablation}) with standard deviation. For Table~\ref{tab:compact_graph_comparison}, standard deviations are estimated via bootstrap resampling. The exact prompts used at each stage are provided in the Appendix~\ref{sec:prompts},~\ref{sec:qa_prompts}.

\subsection{Dataset Statistics}
The test splits of HotpotQA and MuSiQue consist of 1000 samples each. We provide the QA evaluation results for Wikontic and HippoRAG for the whole evaluation set, and for AriGraph, we report the openly available results for 200 test samples. To compare the statistics of KGs, we use 80 MuSiQue samples that are commonly available for all compared methods. 

\subsection{Computational resources}
All experiments on knowledge graph (KG) construction and question answering (QA) were conducted using the OpenAI and OpenRouter APIs. Across all datasets, models, and ablation studies, KG construction and QA (each QA experiment repeated three times to compute mean and standard deviation) required a total cost of approximately \$500.

\subsection{Prompts for triplet extraction}
\label{sec:prompts}
Here we provide excerpts of prompts that were used in
our KG construction pipeline.  \ref{box:candidate_triplet_extraction} was used for candidate triplet extraction. Subsequently, \ref{box:backbone_refinement_entity} was used to refine entity types for both subject and object entities. \ref{box:backbone_refinement_relation} was used for choosing relations among those that can legally connect chosen entity types by Wikidata constraints. Finally, \ref{box:entity_linking} was used to refine surface forms of subject and object. All prompts, instructions, and in-context examples are available with the code.
\begin{tcolorbox}[
  title=\textbf{Prompt 1. Candidate Triplet Extraction},
  colback=gray!3, colframe=gray!60, coltitle=black,
  fonttitle=\bfseries,
  left=2pt, right=2pt, top=2pt, bottom=2pt, 
  boxsep=2pt
]
\refstepcounter{colorboxctr}
\label{box:candidate_triplet_extraction}
\small
You are an algorithm designed to extract structured knowledge from texts to build a Wikidata-like knowledge graph.

A knowledge graph consists of \textbf{triplets} in the format
\texttt{(subject, relation, object)}, where:

\begin{itemize}[leftmargin=1.2em, itemsep=1pt, topsep=2pt]
  \item Subject: A named entity or concept describing
  a group of people, events, or abstract objects that serves as 
  the source of the relation.
  \item Relation: A Wikidata-style predicate connecting
  the subject and the object.
  \item Object: A named entity or concept describing
  a group of people, events, or abstract objects related to the subject.
\end{itemize}

Additionally, some triplets may have \textbf{qualifiers} that provide more context (e.g., date, place, or other attributes). Qualifiers should have relations and objects like triplets do, but instead of a subject, their relation connects an object and the triplet they qualify. Qualifiers must always be attached to a triplet and never exist as standalone triplets. \\

You will receive a text labeled \textbf{``Text:''}.
Your task is to extract meaningful triplets that represent
factual relationships. \\

\textbf{Output Format.} Return only triplets in
\textbf{JSON format} as a list of dictionaries:
\begin{itemize}[leftmargin=1.2em, itemsep=1pt, topsep=2pt]
  \item \texttt{"subject"}: Subject entity.
  \item \texttt{"relation"}: Relation connecting subject and object.
  \item \texttt{"object"}: Object entity.
  \item \texttt{"qualifiers"}: List of dictionaries, each with:
  \begin{itemize}[leftmargin=1.2em, itemsep=1pt, topsep=2pt]
    \item \texttt{"relation"}: Relation connecting triplet and object.
    \item \texttt{"object"}: Object entity connected to the main triplet.
  \end{itemize}
  \item \texttt{"subject\_type"}: Class that describes the subject.
  \item \texttt{"object\_type"}: Class that describes the object.
\end{itemize}
\end{tcolorbox}


\begin{tcolorbox}[title=Prompt 2. Triplet Backbone Refinement - choosing Relevant Entity Types,
  colback=gray!3, colframe=gray!60, coltitle=black,
  fonttitle=\bfseries,
  left=2pt, right=2pt, top=2pt, bottom=2pt, 
  boxsep=2pt]
\refstepcounter{colorboxctr}
\label{box:backbone_refinement_entity}
\small
You are given a factual triplet extracted from text. 
The triplet follows the format (subject, relation, object), 
where: 
\begin{itemize}
    \item Subject: A named entity or concept that represents a person, group, event, or abstract entity serving as the source of the relation.
    \item Relation: A Wikidata-style predicate that defines the connection between the subject and the object.
    \item Object: A named entity or concept that represents a person, group, event, or abstract entity related to the subject.
    \item Subject type: a class that describes the object.
    \item Object type: a class that describes the subject.

\end{itemize}

The extracted entity types of both subject and object were mapped to a set of similar Wikidata-style entity types based on semantic similarity. \\

\textbf{Your Task:}\\

You will be provided with the following:

\begin{itemize}
    \item Text: The original sentence or passage from which the triplet was extracted.
    \item Extracted Triplet: The factual triplet derived from the text.
    \item Candidate subject types: similar entity types for subject type of extracted triplet retrieved from Wikidata.
    \item Candidate object types: similar entity types for object type of extracted triplet retrieved from Wikidata.

\end{itemize}

Select the most appropriate candidate entity types for both 
subject and object from the provided candidates that
best match the meaning of previously extracted triplet
and original text.  \\

Provide ONLY an answer in JSON format with 
the following keys:
\begin{itemize}
    \item "subject\_type": Selected subject type candidate.
    \item "object\_type": Selected object type candidate.
\end{itemize}

\end{tcolorbox}


\begin{tcolorbox}[title=Prompt 3. Triplet Backbone Refinement - choosing Relevant Relation,  colback=gray!3, colframe=gray!60, coltitle=black,
  fonttitle=\bfseries,
  left=2pt, right=2pt, top=2pt, bottom=2pt, 
  boxsep=2pt]

\refstepcounter{colorboxctr}
\label{box:backbone_refinement_relation}
\small
You are given a factual triplet extracted from text. 
The triplet follows the format (subject, relation, object), 
where: \\

\begin{itemize}
    \item Subject: A named entity or concept that represents a person, group, event, or abstract entity serving as the source of the relation.
    \item Relation: A Wikidata-style predicate that defines the connection between the subject and the object.
    \item Object: A named entity or concept that represents a person, group, event, or abstract entity related to the subject.
    \item Subject type: a class that describes the object. 
    \item Object type: a class that describes the subject. 
\end{itemize}

The extracted relation has been mapped to a set of similar Wikidata-style relations based on semantic similarity and the entity types they can connect. \\

\textbf{Your Task:}\\

You will be provided with the following: 

\begin{itemize}
    \item Text: The original sentence or passage from which the triplet was extracted. 
    \item Extracted Triplet: The factual triplet derived from the text. 
    \item Candidate relations: list of relation (or in other words property) names similar to the extracted relation from triplet retrieved from Wikidata.
    \item Candidate relations: list of relation (or in other words property) names similar to the extracted relation from triplet retrieved from Wikidata.
\end{itemize}

Select the most appropriate relation candidate from
the provided candidate triplets that best match the meaning of previously extracted triplet and original text. \\

Provide only an answer in JSON format with the following keys:

\begin{itemize}
    \item "relation": Relation for the selected triplet.
 
\end{itemize}

\end{tcolorbox}


\begin{tcolorbox}[title=Prompt 4. Entity Names Refinement,  colback=gray!3, colframe=gray!60, coltitle=black,
  fonttitle=\bfseries,
  left=2pt, right=2pt, top=2pt, bottom=2pt, 
  boxsep=2pt]
\refstepcounter{colorboxctr}
\label{box:entity_name_refinement}
\small
In the previous step, there was extracted a triplet
akin to one in Wikidata knowledge graph from the text.
Triplet contains two entities (subject and object) 
and one relation that connects these subject and object.
Using semantic similarity, we linked subject name with top similar exact names from the knowledge graph built from 
previously seen texts. \\

You will be provided with the following: \\

\begin{itemize}
    \item Text: The original sentence or passage from which the triplet was extracted.
    \item Extracted Triplet: A structured representation in the format { "subject": "...", "relation": "...", "object": "..." }.
    \item Original Subject: A subject name that needs refinement.
    \item Candidate Subjects: A list of possible entity names from previously seen texts. 

\end{itemize}

\textbf{Your Task:} \\

Select the most contextually appropriate subject name from  the Candidate Subjects list that best matches subject from extracted triplet and context of the given Text.
\begin{itemize}
    \item If an exact or semantically appropriate match is found, return the corresponding name exactly as it appears in the list.
    \item If no suitable match exists, return the string "None".
    \item Do not modify name from the candidate list in case of 
match, add explanations, or provide any additional text.
\end{itemize}
\end{tcolorbox}


\subsection{Prompts for question answering}
\label{sec:qa_prompts}
Here, we provide excerpts of prompts that were used for KG grounded question answering. \ref{box:entity_extraction} was used to extract entities relevant to the question. Then, with \ref{box:entity_linking} the LLM was instructed to choose relevant entities among entities similar to the extracted ones in the KG. \ref{box:question_decomposition} was used to decompose the question on a single-hop subquestion conditioned on extracted entities or previously answered subquestions. \ref{box:if_answered} was used to check if a question is answered by a sequence of subquestions and corresponding answers. All prompts, instructions, and in-context examples are available with the code.

\begin{tcolorbox}[title=Propmt 5. Entity extraction for question answering,  colback=gray!3, colframe=gray!60, coltitle=black,
  fonttitle=\bfseries,
  left=2pt, right=2pt, top=2pt, bottom=2pt, 
  boxsep=2pt]
\refstepcounter{colorboxctr}
\label{box:entity_extraction}
\small
Extract wikidata-like entities from the question below. It is guaranteed that there is at least one mentioned entity. \\

Extract any entity, whether name entity or an abstract entity, that might help retrieve the information to answer the question. \\

Provide output in json format, no additional symbols. Output should be represented as a LIST of extracted entities' names. \\

\end{tcolorbox}


\begin{tcolorbox}[title=Prompt 6. Entity linking for question answering,  colback=gray!3, colframe=gray!60, coltitle=black,
  fonttitle=\bfseries,
  left=2pt, right=2pt, top=2pt, bottom=2pt, 
  boxsep=2pt]
  
\refstepcounter{colorboxctr}
\label{box:entity_linking}
\small
Task: Identify relevant entities from a pre-constructed knowledge graph that might help to answer a provided question. \\

Input Structure:
\begin{itemize}
    \item The question will be labeled as "Question:".
    \item A list of entities from the knowledge graph will be labeled as "Entities:".
\end{itemize}

Selection Criteria:
\begin{itemize}
    \item Relevance means an entity is directly or indirectly useful for answering the question.
    Look for names, events, dates, and other related concepts or entities that match or connect to key concepts in the question.
    \item     Do not ignore possible indirect relevance (e.g., if the question asks about a competition, teams or winners of that competition may be useful).

\end{itemize}

Response Format:
\begin{itemize}
    \item Always return at least one relevant entity. It is guaranteed that there is at least one.
    \item The output must be a JSON list of dictionaries, where each dictionary contains a key "entity": the name of the chosen relevant entity
    \item  Do not return an empty list. Select the best possible options.
\end{itemize}
\end{tcolorbox}


\begin{tcolorbox}[title=Prompt 7. Question decomposition,  colback=gray!3, colframe=gray!60, coltitle=black,
  fonttitle=\bfseries,
  left=2pt, right=2pt, top=2pt, bottom=2pt, 
  boxsep=2pt]

\refstepcounter{colorboxctr}
\label{box:question_decomposition}
\small

You are an assistant for stepwise question decomposition. \\

You will be given three inputs: 

\begin{itemize}
    \item An original multi-hop question.
    \item  A 1-hop sub-question that has already been answered.
    \item The answer to that 1-hop sub-question.
\end{itemize}

Your task:

\quad Reformulate the original multi-hop question by integrating obtained answer from sub-question, so the new question has (n-1) hops. \\

Rules:
\begin{itemize}
    \item Only perform one reasoning hop at a time. Do not generate additional reasoning steps beyond this hop.
    \item Do not include explanations or text, just reformulated question.
\end{itemize}

\end{tcolorbox}

\begin{tcolorbox}[title=Prompt 8. Check if a question is answered,  colback=gray!3, colframe=gray!60, coltitle=black,
  fonttitle=\bfseries,
  left=2pt, right=2pt, top=2pt, bottom=2pt, 
  boxsep=2pt]
\refstepcounter{colorboxctr}
\label{box:if_answered}
\small

You are a reasoning assistant for multi-hop question answering. \\

Your task: Decide whether a list of subquestions and their answers fully resolves the original multi-hop question. \\

Input format:
\begin{itemize}
    \item Original multi-hop question: <text>
    \item Question->answer sequence: [a list of subquestions and their answers, ending with the most recent one]

\end{itemize}

Output rules:
\begin{itemize}
    \item If the sequence of subquestions and answers completely and directly resolves the original multi-hop question, output only the final answer to the original multi-hop question (not just the last subanswer, i.e. answer the original question).
    \item If the sequence is not sufficient and more reasoning or hops are needed, output exactly: NOT FINAL
\end{itemize}

Do not include any prefixes like "Final answer:", "Answer:", suffixes, formatting, original questions or explanations. \\

Output must be a single line: either string with the final answer to the original multi-hop question or the exact string NOT FINAL. \\

<example> \\

\quad Original multi-hop question: Who was the spouse of the person who wrote The Iron Heel? \\

\quad Question->answer sequence: \\

\quad \quad Who wrote The Iron Heel? → Jack London \\

\quad \quad Who was the spouse of Jack London? → Charmian London \\

\quad Expected output: \\

\quad \quad Charmian London \\
</example> \\

<example> \\

\quad Original multi-hop question: Which country’s capital is closest to the birthplace of Nikola Tesla? \\

\quad Question->answer sequence: \\

\quad \quad Where was Nikola Tesla born? → Smiljan, Croatia \\

Expected output: \\

\quad \quad NOT FINAL \\
</example>

\end{tcolorbox}


\subsection{Triplet extraction pipeline examples}
\label{sec:ie_examples}

\begin{figure*}[t]
    \includegraphics[width=0.95\textwidth]{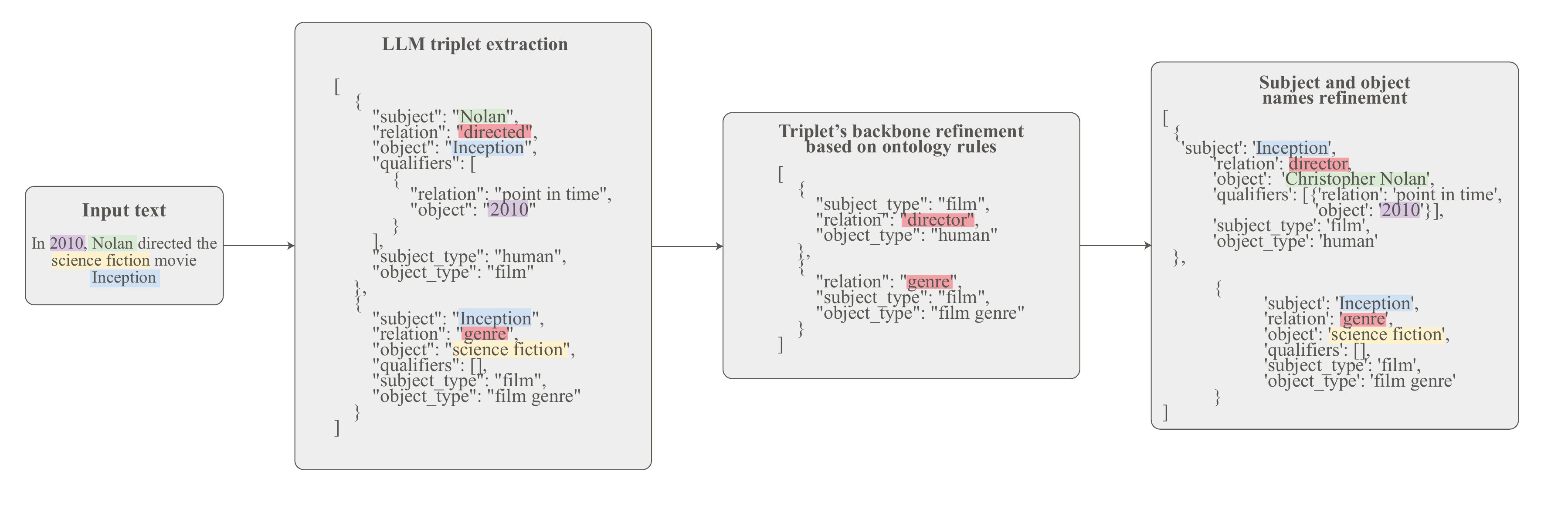}
  \caption{Overview of the multi-stage pipeline for KG extraction from unstructured text. The process consists of (1) LLM-based triplet extraction, (2) ontology-based validation of triplet structure, and (3) entity linking and normalization.}
  \label{fig:overall_details}
\end{figure*}

\begin{figure*}
    \includegraphics[width=\textwidth]{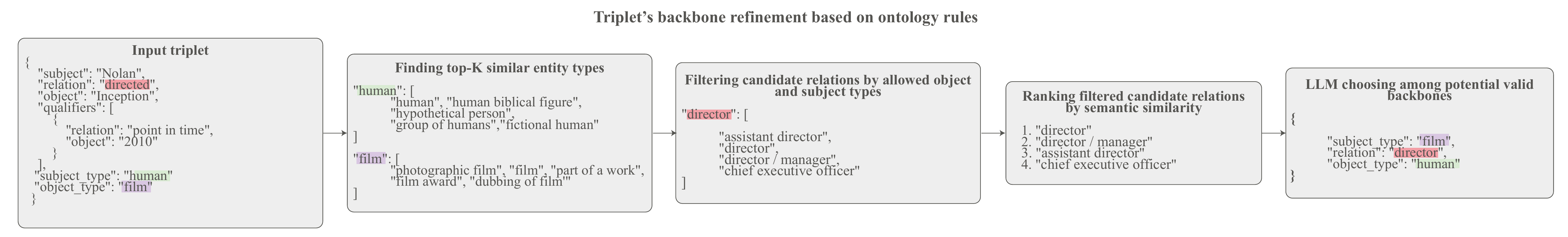}
  \caption{Ontology-based triplet refinement process. For each extracted triplet, we retrieve and extend candidate entity types using Wikidata’s type hierarchy, identify valid relations allowed to use between extracted entities based on ontology constraints, and re-rank relation candidates using semantic similarity. The final triplet configuration is selected by an LLM.}
  \label{fig:backbone}
\end{figure*}

\begin{figure*}
  \includegraphics[width=\textwidth]{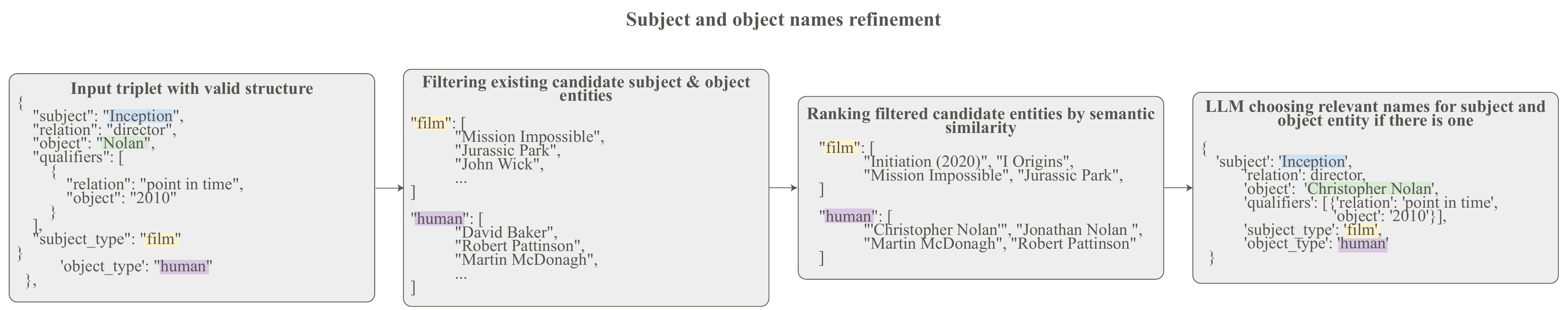}
  \caption{Entity refinement step for KG construction. For each refined triplet, candidate subject and object entities are retrieved from the existing KG based on their type and semantic similarity. An LLM determines whether the extracted entity matches an existing one or should be preserved as a new entry. This process reduces redundancy and supports incremental KG updates.}
  \label{fig:names}
\end{figure*}

\end{document}